\documentclass[pmlr]{jmlr}

\usepackage{booktabs}
\usepackage[load-configurations=version-1]{siunitx} 
\usepackage{outlines}
\usepackage{float}

\makeatletter
\def\set@curr@file#1{\def\@curr@file{#1}} 
\makeatother

\theorembodyfont{\upshape}
\theoremheaderfont{\scshape}
\theorempostheader{:}
\theoremsep{\newline}

\jmlrvolume{182}
\jmlryear{2022}
\jmlrworkshop{Machine Learning for Healthcare}

\title[Data Augmentation for Robust Electrocardiogram Prediction]{GeoECG: Data Augmentation via Wasserstein Geodesic Perturbation for Robust Electrocardiogram Prediction} 

\author{\Name{Jiacheng Zhu$^*$} \Email{jzhu4@andrew.cmu.edu} \\
\addr Department of Mechanical Engineering, Carnegie Mellon University\\
Pittsburgh, PA 15213, USA
\AND
\Name{Jielin Qiu$^*$} \Email{jielinq@andrew.cmu.edu}  \\
\addr Computer Science Department, Carnegie Mellon Universityn\\
Pittsburgh, PA 15213, USA
\AND
\Name{Zhuolin Yang} \Email{zhuolin5@illinois.edu} \\
\addr  Computer Science Department, University of Illinois at Urbana-Champaign\\
Urbana, IL 61801, USA
\AND
\Name{Douglas Weber} \Email{dougweber@cmu.edu} \\
\addr Department of Mechanical Engineering, Carnegie Mellon University\\
Pittsburgh, PA 15213, USA
\AND
\Name{Michael A. Rosenberg} \Email{michael.a.rosenberg@cuanschutz.edu} \\
\addr University of Colorado Denver - Anschutz Medical Campus\\
Aurora, CO 80045, USA
\AND
\Name{Emerson Liu} \Email{emersonliu@msn.com} \\
\addr Allegheny Health Network \\
Pittsburgh, PA 15212, USA
\AND
\Name{Bo Li} \Email{lbo@illinois.edu} \\
\addr Computer Science Department, University of Illinois at Urbana-Champaign\\
Urbana, IL 61801, USA
\AND
\Name{Ding Zhao} \Email{dingzhao@cmu.edu} \\
\addr Department of Mechanical Engineering, Carnegie Mellon University\\
Pittsburgh, PA 15213, USA
\AND
\footnotemark[1] \addr {\normalfont \footnotesize authors of equal contribution}
}

\begin{document}

\maketitle

\begin{abstract}
There has been an increased interest in applying deep neural networks to automatically interpret and analyze the 12-lead electrocardiogram (ECG). 
The current paradigms with machine learning methods are often limited by the amount of labeled data. This phenomenon is particularly problematic for clinically-relevant data, where labeling at scale can be time-consuming and costly in terms of the specialized expertise and human effort required. 
Moreover, deep learning classifiers may be vulnerable to adversarial examples and perturbations, which could have catastrophic consequences, for example, when applied in the context of medical treatment, clinical trials, or insurance claims.
In this paper, we propose a physiologically-inspired data augmentation method to improve performance and increase the robustness of heart disease detection based on ECG signals.
We obtain augmented samples by perturbing the data distribution towards other classes along the geodesic in Wasserstein space. 
To better utilize domain-specific knowledge, we design a ground metric that recognizes the difference between ECG signals based on physiologically determined features.
Learning from 12-lead ECG signals, our model is able to distinguish five categories of cardiac conditions. Our results demonstrate improvements in accuracy and robustness, reflecting the effectiveness of our data augmentation method. 
\end{abstract}

\section{Introduction}
Heart and cardiovascular diseases are the leading global cause of death, with 80\% of cardiovascular disease-related deaths due to heart attacks and strokes. The 12-lead ECG can be considered as the foundation of cardiology and electrophysiology. It provides unique information about the structure and electrical activity of the heart as well as systemic conditions, through changes in timing and morphology of the recorded waveforms. Consequently, the clinical 12-lead ECG when correctly interpreted, remains a primary tool for detecting cardiac abnormalities and screening at-risk populations for heart-related issues.

Accurate ECG interpretations of acute cardiac conditions are critical for timely, efficient, and cost-effective interventions. Consequently, achievement of reliable machine assisted ECG interpretations could significantly impact patient outcomes~\citep{zhu2022physiomtl}.   
With the development of machine learning and deep learning methods, it may also be possible to identify additional previously unrecognized signatures of disease. Many methods have been explored for diagnosing physiological signals, i.e.,  EEG, ECG, EMG, etc \citep{Liu2019MultimodalER,Shanmugam2019MultipleIL,CtAllard2019DeepLF}. 
Due to limited data and sensitive modeling frameworks, the diagnostic performance of developed algorithms is not always robust. Also, it has been shown that deep learning models for ECG data could be susceptible to adversarial attacks. \citep{han2020deep,hossain2021_ecg_adv_gan,chen2020_aaai_ecg_adv}.

To tackle the problem caused by \textit{adversarial data distributions}, people have proposed both empirical and certified robust learning approaches, such as adversarial training \citep{madry2017towards_emp_adv} and certified defense approaches \citep{cohen2019certified_cert_rob,li2020sok_SOK,li2021tss_TSS}. 
It has already been shown that \textit{data augmentation} strategies \citep{Rebuffi2021DataAC,Rebuffi2021FixingDA,Gao2020FuzzTB,volpi2018generalizing_data_aug_dro,Ng2020SSMBASM} or more training data \citep{carmon2019unlabeled_imp_rob} can improve the performance and increase the robustness of deep learning models. Specifically, augmenting data with random Gaussian noise \citep{cohen2019certified_cert_rob} or transformations \citep{li2021tss_TSS} yields certifiable smoothed models. Mixup methods \citep{Zhang2018mixupBE,greenewald2021_kmixup}, which augment data with weighted averages of training points, also promote the certifiable robustness \citep{jeong2021smoothmix}. However, different types of data usually contain critical domain-specific properties. 

While people have effectively applied different neural network architectures to ECG classification problems, it has become an increasing concern that these neural networks are susceptible to adversarial examples~\citep{han2020deep}.
Several studies~\citep{raghu22a,nonaka2020electrocardiogram_ecg_data_aug} have explored data augmentation techniques for ECG datasets. Nevertheless, unlike other fields such as computer vision~\citep{Zhao2020MaximumEntropyAD} or NLP~\citep{morris2020textattack_data_aug_adv_nlp}, the effect of data augmentation on ECG deep learning robustness is less explored.



In this paper, we propose a data augmentation method from a probability and geometric perspective. Following the notion of optimal mass transportation theories~\citep{villani2009optimal}, we perturb the data distributions along the geodesic in a Wasserstein space. Specifically, the ground metric of this Wasserstein space is computed via comparing the geometry of ECG signals, which exploits the cardiovascular properties. 
In summary, our contribution is threefold:
\begin{enumerate}
    \item Our proposed data augmentation method augments samples by perturbing an empirical distribution towards samples of other classes. This data augmentation scheme can preserve the local structure on the data manifold.  
    \item To perform the computation of Wasserstein barycenters, we propose a similarity metric for ECG signals by comparing their shapes, where we consider each beat of an ECG as a \textit{continuous probability} and compute the corresponding Wasserstein distance.
    \item We validate our method on the PTB-XL dataset, which covers a variety of conditions, including Myocardial Infarction (MI), ST/T Change (STTC), Conduction Disturbance (CD), and Hypertrophy (HYP), collected from subjects of different ages and gender. We compare our methods with a list of baseline methods in terms of the standard prediction performances and robust prediction under adversarial attacks. 
\end{enumerate}

\subsection*{Generalizable Insights about Machine Learning in the Context of Healthcare}

ECG signals can be treated as continuous sequential data and directly fed into deep learning models, since some neural networks architectures, such as ResNet, can effectively capture information from raw ECG signals. However, our study emphasizes the potential importance of utilizing physiologically informed features that are intrinsically encoded in the structure or shape of the ECG waveforms. This statement is motivated by the following: (1) Specific components of ECG signals are generated from different parts of the cardiac cycle and represent different physiology (2) The periodicity (or absence) of expected ECG waveforms contains valuable information beyond that contained within the waveforms themselves. (3) Human experts specify ECG categories predominantly based on coarser visual features (shape and morphology) that can reflect a broad variety of structural or conduction abnormalities. (4) The advantages of incorporating some prior knowledge into models is supported by the fact that the leading method of the 2020 PhysioNet challenge only used handcrafted features rather than raw signals. 

Hence, we suggest exploring and developing algorithms based on electrocardiograms' properties and physiological features. While decompiling ECG signals into individual wave components (P, QRS, T waves) is challenging, our method of comparing ECG beats with respect to their underlying geometry offers a principle approach to discriminating ECG signals. 

\section{Related Work}
\paragraph{ECG Robustness} The robustness of ECG has recently drawn more attention. \citet{Venton2021RobustnessOC} generated clean and noisy ECG datasets to test the robustness of different models. \citet{Hossain2021ECGATKGANRA} proposed Conditional GAN, which claimed to be robust against adversarial attacked ECG signals. \citet{Venton2021InvestigatingTR} explored the impact of different physiological noise types and differing signal-to-noise ratios (SNRs) of noise on  ECG classification performance.

\paragraph{Deep learning in ECG} Deep learning approaches have been rapidly adopted across a wide range of fields due to their accuracy and flexibility but require large labeled training sets.
With the development in machine learning, many models have been applied to ECG disease detection \citep{Kiranyaz2015ConvolutionalNN,pmlr-v149-nonaka21a,Khurshid2021ElectrocardiogrambasedDL,Raghunath2021DeepNN,Giudicessi2021ArtificialIA,Strodthoff2021DeepLF}. Al-Zaiti et al. predicted acute myocardial ischemia in patients with chest pain with a fusion voting method \citep{AlZaiti2020MachineLP}. Acharya et al. proposed a nine-layer deep convolutional neural network (CNN) to classify heartbeats in the MIT-BIH Arrhythmia database \citep{Acharya2017ADC,Moody2001TheIO}. Shanmugam et al. estimate a patient’s risk of cardiovascular death after an acute coronary syndrome by a multiple instance learning framework \citep{Shanmugam2019MultipleIL}. Recently, Smigiel et al. proposed models based on SincNet \citep{Ravanelli2018SpeakerRF} and used entropy-based features for cardiovascular diseases classification \citep{smigiel2021ecg}. The transformer model has also recently been adopted in several ECG applications, i.e., arrhythmia classification,  abnormalities detection, stress detection, etc  \citep{Yan2019FusingTM,Che2021ConstrainedTN,Natarajan2020AWA,Behinaein2021ATA,Song2021TransformerbasedSF,Weimann2021TransferLF}.

\paragraph{Data augmentation for ECG} The data augmentation task has also been explored for ECG applications in the previous studies. \citet{Martin2021RealtimeFS} tried to use oversampling method to augment the imbalanced data. \citet{ClementVirgeniya2021AND} tried to feed the data into the adaptive synthetic (ADASYN) \citep{He2008ADASYNAS} based sampling model, which utilized a weighted distribution for different minority class samples depending upon the learning stages of difficulty, instead of using synthetic models such as synthetic minority oversampling technique (SMOTE). \citet{Liu2021MultiLabelCO} augmented the ECG data by using a band-pass filter, noise addition, time-frequency transformation, and data selection. Data augmentation is also a good method to deal with imbalanced ECG dataset~\citep{Qiu2022OptimalTB}. 
Recently, a task-dependent learnable data augmentation policy~\citep{raghu22a} has been developed for 12-lead ECG detection. This study showed that data augmentation techniques are not always helpful. 

\paragraph{Data augmentation \& robustness in ML: }
A promising way to enable robust learning is to provide adversarially perturbed samples with data augmentation. 
It has already been shown that data augmentation \citep{rebuffi2021data_aug_rob_deepmind,volpi2018generalizing_data_aug_dro} or more training data \citep{carmon2019unlabeled_imp_rob,deng2021improving_ood_data_rob_gaussian_mix} can improve the performance and increase the robustness of deep learning models.

\citet{Zhang2018mixupBE}  proposed Mixup,  an effective model regularizer for data augmentation that encourages linear interpolation in-between training examples, which has been applied in sequential data.
\citet{Zhang2020SeqMixAA} augmented the queried samples by generating extra labeled sequences. \citet{Guo2020SequencelevelMS}  created new synthetic examples by softly combining input/output sequences from the training set. \citet{Guo2020NonlinearMO} embraced a nonlinear interpolation policy for both the input and label pairs, where the mixing policy for the labels is adaptively learned based on the mixed input. 
Perhaps there is one recent work that is conceptually close to our study where a k-mixup data augmentation~\citep{greenewald2021_kmixup}, guided by Optimal Transport (OT), is proposed to improve the generalization and robustness of neural networks. This method uses optimal coupling to interpolate for vicinal data samples that respect local structures. Our method also enjoys this benefit since the Wasserstein barycenter also exploits local distribution structures. However, this study uses $l2$ cost as the ground metric, which could be ineffective when dealing with high-dimensional data. On the contrary, our study utilizes a ground metric that compares ECG signals according to their cardiovascular characteristics.


\begin{figure}[t]
\centering
\includegraphics[width=0.99\textwidth]{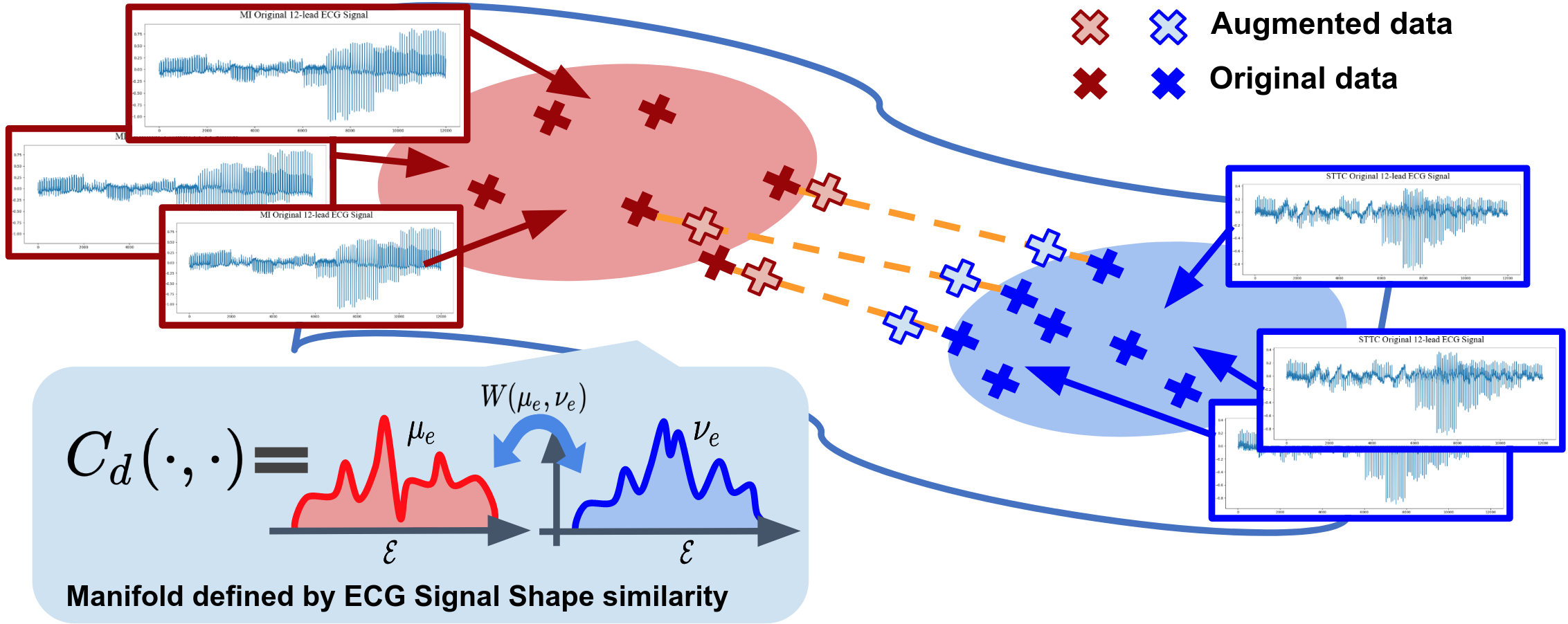}
\caption{
Our data augmentation creates perturbed samples toward vicinal other-class samples. The perturbation lies on the geodesic connecting two distributions on a Wasserstein space, whose ground cost metric is computed via another level Wasserstein distance that compares the geometric shape of ECG signals.
}
\label{fig:framework}
\end{figure}

\section{Methods}\label{sec_OT}


In this work, we focus on the model's performance on adversarial examples, which are generated by adding imperceptible noises on clean inputs to mislead ML models' predictions through well-designed attack algorithms~\citep{szegedy2013intriguing, goodfellow2014explaining, eykholt2018robust}. Given such malicious scenarios and ML security considerations, robust deep learning has been studied extensively~\citep{salman2019provably, allen2022feature}, to develop effective learning algorithms to build robust models.

\subsection{Robust Deep Learning with Data Augmentation}

It is imperative to obtain a deep learning model that is operational in the presence of potentially adversarial shifts in data distribution. A common way to describe this procedure is through the framework of distributional robust optimization \citep{weber2022certifying_cert_ood_gen_dro}. Specifically, denote $P$ as the joint data distribution over features $X \in \mathcal{X}$ and labels $Y \in \mathcal{Y}$, and let $h_\theta : \mathcal{X} \mapsto \mathcal{Y}$ be a family of predictive function parameterized by $\theta$. Given a loss function $l : \mathcal{Y} \times \mathcal{Y} \mapsto \mathbb{R}$, we turn to solve the following optimization problem:
\begin{equation}
    \min_{\theta} \sup_{Q \in \mathcal{U}_P} \mathbb{E}_{(X, Y) \sim Q} [l(h_{\theta}(X), Y)],
\end{equation}
where $\mathcal{U}_P \subseteq \mathcal{P(Z)} $ is a set of probability distribution. Intuitively, this objective aims at finding the worst-case optimal predictor $h^*_\theta$ when the data distribution $P$ is perturbed towards some distribution $\mathcal{U}_P$.

In this work, we follow the distribution perturbing adversary frameworks~\citep{sinha2017certifying_distribution_rob,mehrabi2021fundamental_distributionally_adv} 
wherein the adversarial distributions can be viewed as the neighbor distribution of clean data distribution, characterized by certain distribution distance metrics (e.g., Wasserstein distance~\citep{villani2009optimal}). It is challenging to access this adversarial distribution explicitly. Thus, inspired by recent studies~\citep{carmon2019unlabeled_imp_rob,zhai2019adversarially_rob_gaussian,dan2020sharp_rob_gaussian}, we make further assumptions for the distribution of the data by writing out the joint data distribution $P(X, Y) $ as the product of conditional distributions $P(X, Y) = P(X|Y)P(Y)$. Since we focus on the multi-classification task, we denote $P_k(X) = P(X | Y_k)$ as the data distribution of one class $k$. 
During the data augmentation, we aim to perturb the class $i$'s data distribution $P_i(X)$ towards class $j$'s distribution $P_j(X), i \neq j$ since we believe the data samples lying on the geodesic can be served as adversarial samples.



We can illustrate our intuitions as follows:
\textbf{(1)} Instead of the datapoint-specific adversarial perturbations that are aimed to attack one specific sample, the directed augmented data distribution can be considered as universal perturbations~\citep{moosavi2017robustness_boundary} that cause label change for a set of samples from the perturbed distribution $\mathcal{U}_P$.
\textbf{(2)} Such perturbation matches the global manifold structure of the dataset~\citep{greenewald2021_kmixup}, therefore promoting a smoother decision boundary.
\textbf{(3)} It is shown in~\citet{wei2020theoretical_expansion} that this augmentation strategy improves the expansion of the neighborhood of class-conditional distributions.  
More significantly, this formulation allows us to employ the results from OT theories~\citep{villani2009optimal} and Wasserstein Barycenter~\citep{agueh2011barycenters} thus firmly estimating the perturbed distribution $\mathcal{U}_P$.

\subsection{Data Augmentation by Perturbation on the Geodesic}\label{sec:DA}

Let $\mathcal{X}$ be an arbitrary space. Assume $d(\cdot, \cdot) : \mathcal{X} \times \mathcal{Y} \mapsto \mathbb{R}^+$ is the ground metric cost function.
The well-known Wasserstein distance originated from the Optimal Transport (OT) problem which aims at finding an optimal coupling $\pi$ that minimizes the transportation cost. 
\begin{definition}
(Wasserstein Distances). For $p \in [1, \infty]$ and probability measures $\mu$ and $ \nu \in \mathcal{M}(\mathcal{X})$. The $p-$Wasserstein distance between them is defined as
\begin{equation}
    W_p(\mu, \nu) := \left( \inf_{\pi \in \Pi} \int_{\mathcal{X} \times \mathcal{X}} d^p(x, y) d \pi(x, y) \right)^{1/p} , \text{  } (x, y) \in \mathcal{X} \times \mathcal{X}
    \label{eq:w-dist}
\end{equation}
where $\Pi$ is the set of all probability measures on $\mathcal{X} \times \mathcal{X}$.
\end{definition}
Considering the path of distributions (a geodesic) $p_t$ that interpolates between two distributions $\mu$ and $\nu$, one of the most intriguing properties of this interpolation is that it will preserve the basic structure of $\mu$ and $\nu$. In other words, such perturbation can be viewed as an optimal transport map that pushes forward $\mu$ along the geodesic that connects $\mu$ and $\nu$. 

\begin{definition}
(Geodesics in Wasserstein space). Let $\mu$ and $\nu$ be two distributions. Consider a map $m: [0, 1] \mapsto \mathcal{M}(\mathcal{X})$ taking $[0, 1]$ to the set of distributions, such that $m(0) = \mu$ and $m(1) = \nu$, where $\mathcal{M}(\mathcal{X})$ is the set of Borel measures on $\mathcal{X}$. 
Thus $(p_{\alpha}: 0 \leq \alpha \leq 1)$ is a path connecting $\mu$ and $\nu$, where $p_{\alpha}=m(\alpha)$. The length of $m$ --- denoted by $L(m)$ --- is the supremum of $\sum_{i=1}^{K} W\left(m\left(\alpha_{i-1}\right), m\left(\alpha_{i}\right)\right)$ over all $m$ and all $0=\alpha_{1}<\cdots<\alpha_{K}=1$. 
Therefore, there exists such a path $m$ such that $L(m) = W(\mu, \nu)$ and $(p_{\alpha}: 0 \leq \alpha \leq 1)$ is the geodesic connecting $\mu$ and $\nu$. 
\end{definition}
The definition of the geodesic in Wasserstein space provides us a roadmap to obtain the perturbed distributions, as it boils down to the Wasserstein Barycenter problem.

\begin{definition}
 (Wasserstein Barycenter). The Wasserstein barycenter of a set of measures $\{\nu_1, ..., \nu_N \}$ in a probability space $\mathbb{P} \subset \mathcal{M}(\mathcal{X})$ is a minimizer of objective $f_{wb}$ over $\mathbb{P}$, where
 \begin{equation}
     f_{wb}(\mu) := \frac{1}{N} \sum_{i=1}^N \alpha_i W(\mu, \nu_i),
 \end{equation}
 where $\alpha_i$ are the weights such that $\sum \alpha_i = 1$ and $\alpha_i > 0$. 
\end{definition}
If we are using uniform weights and consider all the distributions, the barycenter is the Fr\'echet mean, or the Wasserstein population mean~\citep{bigot2017geodesic_pca}.
Also, it is known that when we have only two samples, the barycenter corresponds to the interpolation between two distributions along the geodesic.

In this case, given  class-conditional data distributions  $P_i$ and $P_j$, the perturbed augmentation which interpolates along the geodesic can be obtained via 
\begin{equation}
   \Tilde{P}_{ij} =  \inf_{{P}_\alpha} ~ (1 - \alpha)W(P_i, {P}_\alpha) + \alpha W({P}_\alpha,P_j) \text{ where  } \alpha \in (0, \epsilon)
\end{equation}
Then the augmented samples can be obtained by sampling $(\Tilde{x}_i, y_i) \sim \Tilde{P}_{ij}$. Later on, we will provide an algorithmic derivation of this augmentation procedure for discrete data samples. 

\begin{figure}[htp]
\centering
\includegraphics[width=0.99\textwidth]{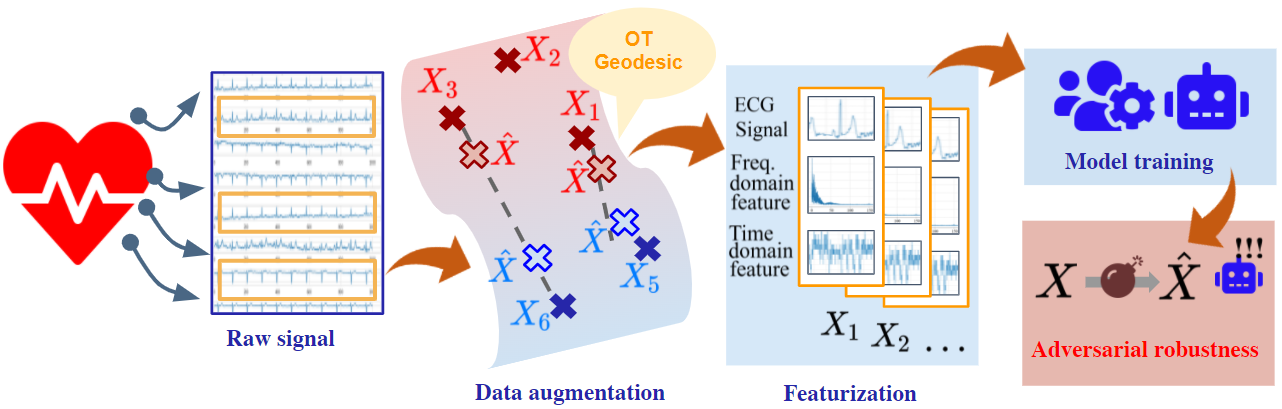}
\caption{The semantic representation of our pipeline. }
\label{fig:physio_pipeline}
\end{figure}

\section{Algorithm \& Computation }
\subsection{Computational Optimal Transport}
In practice, we only observe discrete training samples that represents empirical distribution of $P_i$ and $P_j$. 
Consider  $\mathbf{X}_i = \{\mathbf{x}^i_l\}_{l=1}^{n_i}$ and $\mathbf{X}_j = \{\mathbf{x}^j_l\}_{l=1}^{n_j}$ are two set of features from class $i$  and $j$ respectively. The empirical distributions are written as $\hat{P}_i = \sum_{l=1}^{n_i} p_l^i \delta_{x^i_l}$ and $\hat{P}_j = \sum_{l=1}^{n_j} p_l^j \delta_{x^j_l}$ where $\delta_{x}$ is the Dirac function at location $x \in \Omega$, $p_l^i$ and $p_l^j$ are probability mass associated to the sample. 
Then the Wasserstein distance, or equation (\ref{eq:w-dist}), between empirical measures $\hat{P}_i$ and $\hat{P}_j$ becomes
\begin{equation}
    W(\hat{P}_i, \hat{P}_j)  = \inf_{\pi \in \hat{\Pi_{ij}}} \sum_{l=1,k=1}^{n_i, n_j} c(\mathbf{x}^i_l, \mathbf{x}^j_k) \pi_{l,k},
\end{equation}
where $\hat{\Pi_{ij}}:= \{\pi \in (\mathbb{R}^+)^{n_i \times n_j} | \pi \mathbf{1}_{n_j} = \mathbf{1}_{n_i} /n_i, \pi^\top  \mathbf{1}_{n_i} = \mathbf{1}_{n_j} /n_j \}$ with $\mathbf{1}_n$ a length $n$ vector of ones. $c(x, y)$ is the ground cost function that specifies the actually cost to transport the mass, or probability measure, from position $x$ to $y$.  Most studies use $l_2$ norm as the ground metric as there are a lot of desirable properties. However, here we emphasis that it is not appropriate to compare ECG signals with $l_2$ metrics.



\subsection{A Physiology Inspired Metric for ECG} 
\begin{figure}[htp]
\centering
\begin{minipage}[b]{.45\linewidth}
    \centering
	\includegraphics[width=0.60\textwidth]{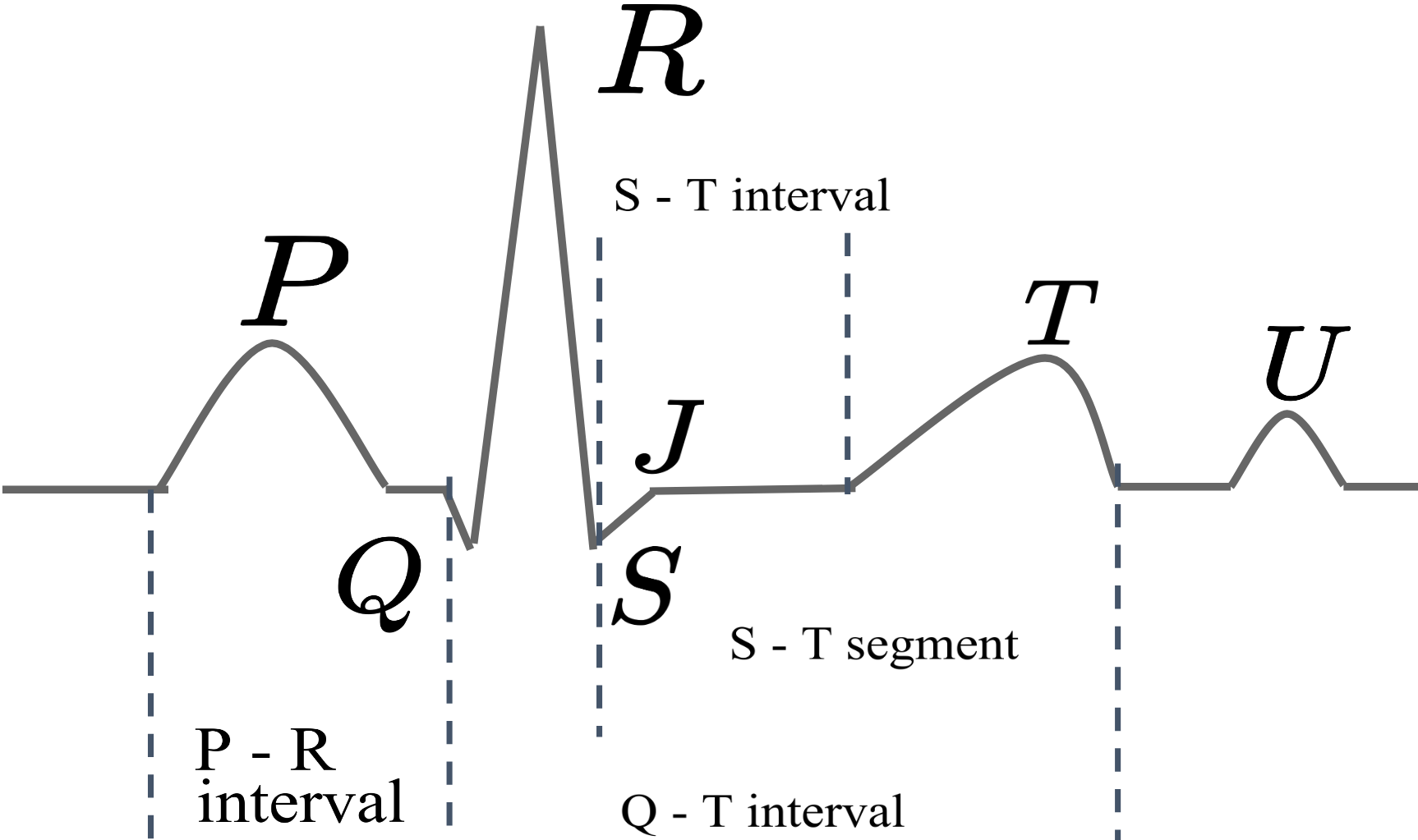}
	\caption{ECG components.}
	\label{fig:ecg_decomp_qrs}
\end{minipage}
\begin{minipage}[b]{.54\linewidth}
	\centering
	\includegraphics[width=0.98\textwidth]{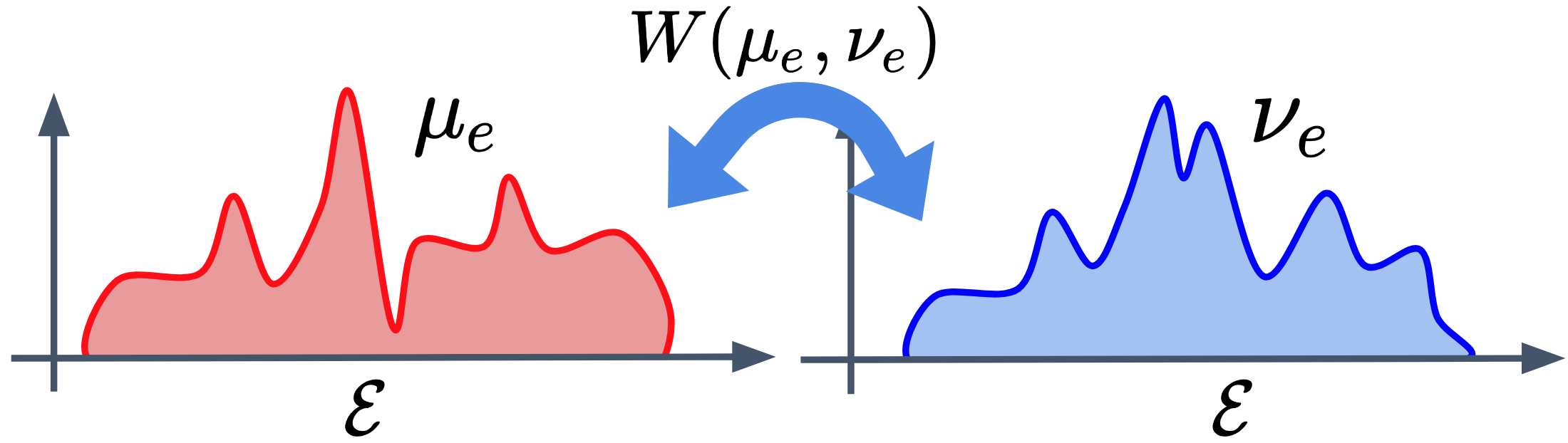}
	\caption{Treat ECG as continuous densities.}
	\label{fig:ecg_shape_w_distance}
\end{minipage}
\end{figure}

In practice, the accurate decomposition~\citep{kanjilal1997fetal_ecg_decomp} of ECG is a crucial step in providing medical diagnosis and services. For example, ventricular heart rate~\citep{kundu2000knowledge_ecg_decomp} is the most common information that is extracted by measuring the time interval between two successive $R$ peaks. While in most computer vision tasks, it is hard to describe features explicitly,  informative characteristics are defined the wave features of the ECG signal, as illustrated in Fig.~\ref{fig:ecg_decomp_qrs}. A great deal of work \citep{zhong2020maternal_decomp_ff,rasti2021aecg_decomp_deep} focused on extracting or decomposing the wave components from ECG signals. However, it is still challenging. 

In this work, we propose to directly compare the shape of two ECGs rather than parsing the ECG into the P wave, QRS, and T wave since it is challenging to process the noisy signals. 
Specifically, we first (1) treat them as probability densities, and then (2) compute a Wasserstein distance between these two densities. Formally, consider two individual ECG beat signals as two density function of time $\mu_e = \mu_e(t)$ and $\nu_e = \nu_e(t)$. The Wasserstein distance is obtained via:
\begin{equation}
    W_e(\mu_e, \nu_e) := \inf_{\pi_e \in \Pi_e} \int_{[0, 1] \times [0, 1]} \| x - y\|_2^2 d \pi_e (x, y),
\end{equation}
where $\Pi_e$ is the joint distribution which has marginals as $\mu_e$ and $\nu_e$. Therefore, now we have a reasonable metric that measures the pairwise similarity between ECG signals $C_d(\cdot, \cdot) = W_e(\cdot, \cdot)$, which can serve as the ground metric in the computation of the Wasserstein barycenter data augmentation procedure.

\paragraph{Computation concerns: batch OT and entropic OT} Discrete optimal transport involves a linear program that has an $O(n^3)$ complexity. Our framework requires the computation of optimal transport in two levels: (1) Use Wasserstein distance to obtain the pairwise similarity of ECG signals. (2) Use Wasserstein Barycenter, which also computes Wasserstein distances, to interpolate between two sets of ECG signal samples from different conditions. Hence, the potential computation issues can not be ignored. 

First of all, we adopted the celebrated entropic optimal transport~\citep{cuturi2013sinkhorn} and used the Sinkhorn algorithm to solve for OT objectives and Barycenters~\citep{janati2020debiased_barycenter}. The Sinkhorn algorithm has a $O(n \log n)$ complexity, thus it can ease the computation burden. In addition, the pairwise Wasserstein distance of ECG signals can be precomputed and stored. Last but not least, we follow the concept of minibatch optimal transport~\citep{fatras2021minibatch_OT} where we sample a batch of ECG samples from each condition during the data augmentation procedure. Whereas minibatch OT could lead to non-optimal couplings, our experimental results have demonstrated that our data augmentation is still satisfactory.

\subsection{Backbone Model} 
\citet{pmlr-v149-nonaka21a} used raw ECG signal as input, however, it is shown that the optimal predictive performance can be achieved  by transformers trained with hand-crafted features \citep{Natarajan2020AWA}. So for the classification model, we take advantage of the transformer encoder \citep{Vaswani2017AttentionIA}, and proposed a Multi-Feature Transformer (MF-Transformer) model. The transformer is based on the attention mechanism \citep{Vaswani2017AttentionIA} and outperforms previous models in accuracy and performance on many tasks~\citep{xu2022prompting,qiu2022mhms}.
The original transformer model is composed of an encoder and a decoder. The encoder maps an input sequence into a latent representation, and the decoder uses the representation along with other inputs to generate a target sequence. Our model is mostly based on the encoder, since we aim at learning the representations of ECG features, instead of decoding it to another sequence.

\begin{figure*}[htp]
	\centering
	\includegraphics[width=0.99\textwidth]{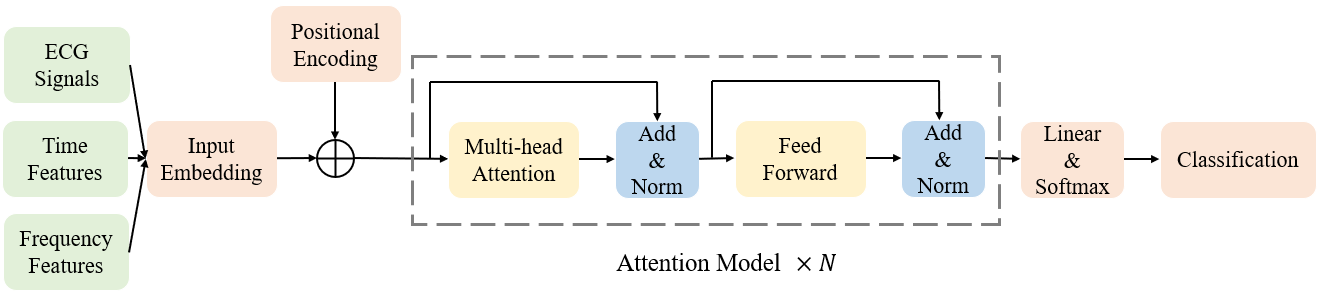}
	\caption{The architecture of the Multi-Feature Transformer model.}
	\label{fig:transformer}
\end{figure*}

As shown in Fig.~\ref{fig:transformer}, the input for the Multi-Feature Transformer is composed of three parts, including ECG raw features, time-domain features, and frequency domain features. First, we feed out the input into an embedding layer and then inject positional information into the embeddings. In our model, the attention model contains $N=5$ same layers, and each layer contains two sub-layers: a multi-head self-attention model and a fully connected feed-forward network. Residual connection and normalization are added in each sub-layer. We use a 1D convolutional and softmax layers for the output to calculate the final output.
More details of the MF-Transformer model is introduced in Appendix~\ref{MF}.

\section{Cohort}
We carried out the experiments on the PTB-XL dataset \citep{Wagner2020PTBXLAL}, which contains clinical 12-lead ECG signals of 10-second length. There are five conditions in total, including Normal ECG (NORM), Myocardial Infarction (MI), ST/T Change (STTC), Conduction Disturbance (CD), and Hypertrophy (HYP). The waveform files are stored in WaveForm DataBase (WFDB) format with 16-bit precision at a resolution of 1$\mu$V/LSB and a sampling frequency of 100Hz.

\subsection{Signal Pre-processing}
\begin{figure}[htp]
\centering
\begin{minipage}[b]{.48\linewidth}
    \centering
	\includegraphics[width=0.98\textwidth]{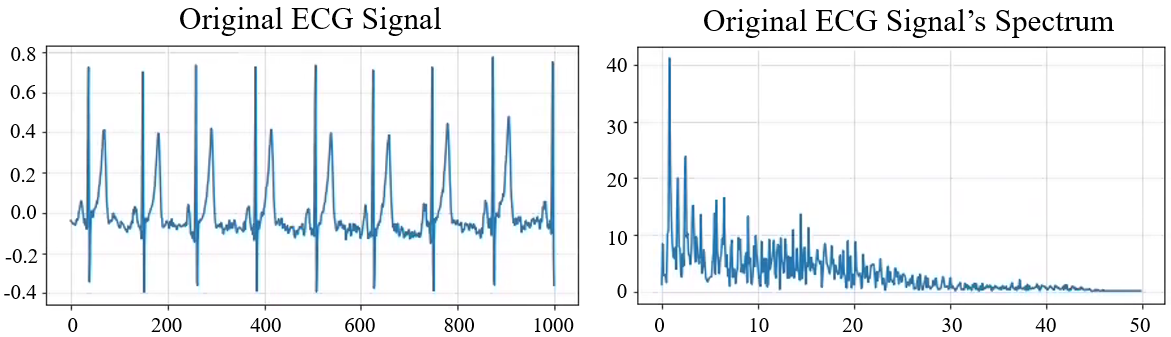}
	\caption{ECG data in the format of time series and spectrum.}
	\label{fig:fft}
\end{minipage}
~
\begin{minipage}[b]{.48\linewidth}
	\centering
	\includegraphics[width=0.98\textwidth]{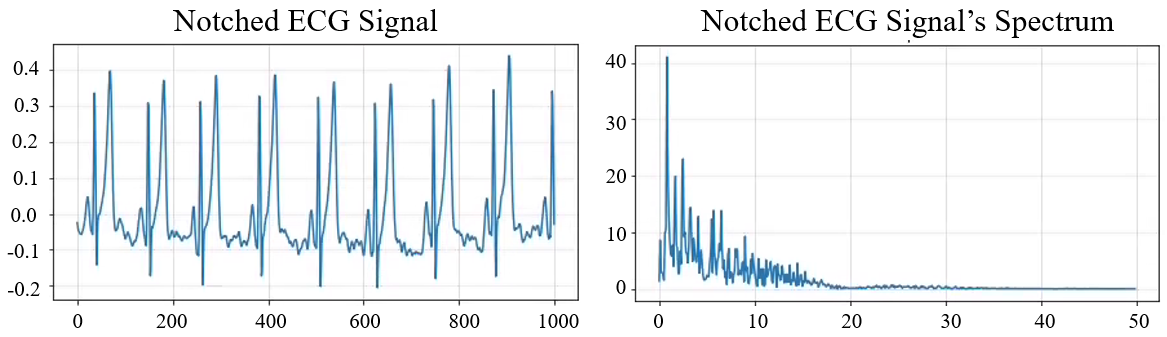}
	\caption{Filtered  ECG data  in the format of time series and spectrum.}
	\label{fig:notch}
\end{minipage}
\end{figure}

First, the raw ECG signals are processed by the wfdb library\footnote{https://pypi.org/project/wfdb/} and Fast Fourier transform (fft) to process the time series data into the spectrum, which is shown in Fig.~\ref{fig:fft}. Then we perform n-points window filtering to filter the noise within the original ECG signals and adopt notch processing to filter power frequency interference (noise frequency: 50Hz, quality factor: 30). An example of the filtered ECG signal result after n-points window filtering and notch processing is shown in Fig.~\ref{fig:notch}.

\begin{figure}[htp]
\centering
\begin{minipage}[b]{.49\linewidth}
	\centering
	\includegraphics[width=0.98\textwidth]{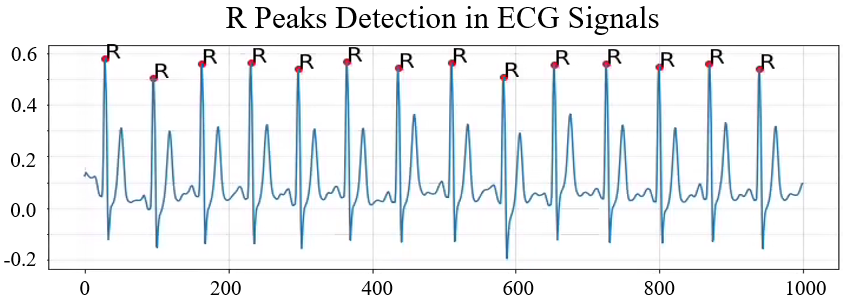}
	\caption{Detecting R peaks in the ECG signals.}
	\label{fig:find_R}
\end{minipage}
\begin{minipage}[b]{.49\linewidth}
	\centering
	\includegraphics[width=0.98\textwidth]{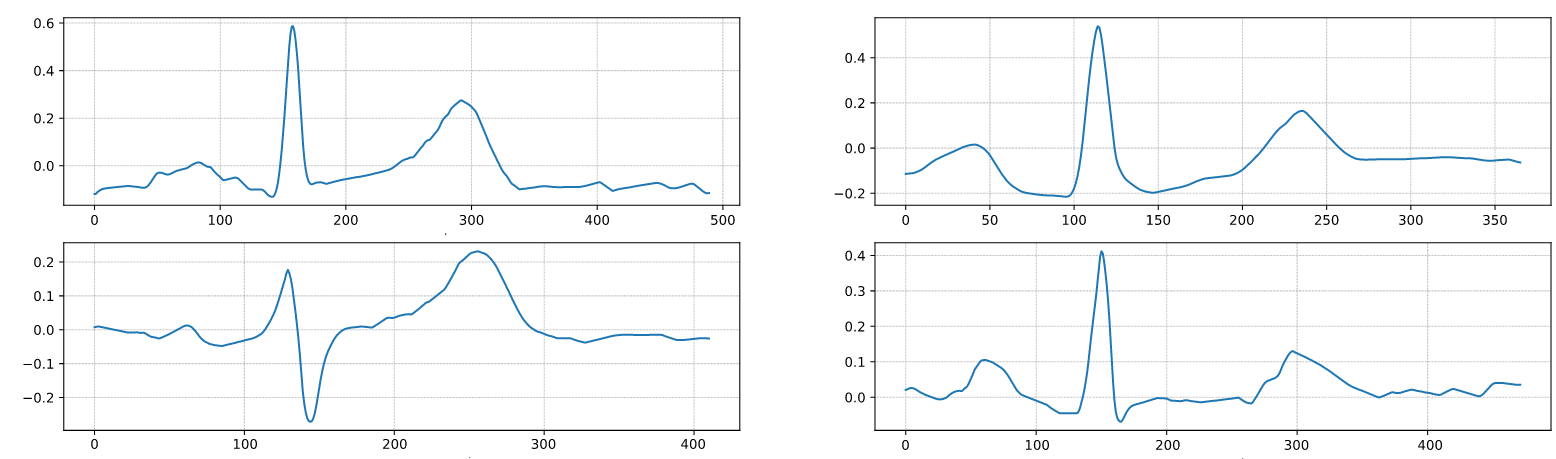}
	\caption{Extracted ECG beats divided by R peaks.}
	\label{fig:divide_R}
\end{minipage}
\end{figure}

Then we perform ECG segmentation by dividing the 10-second ECG signals into individual ECG beats. We first detect the R peaks of each signal by ECG detectors\footnote{https://pypi.org/project/py-ecg-detectors/}, and then slice the signal at a fixed-sized interval on both sides of the R peaks to obtain individual beats. Examples of R peak detection results  and segmented ECG beats are shown in Fig. \ref{fig:find_R} and Fig.~\ref{fig:divide_R}, respectively.

\subsection{Feature Extraction}\label{sec:features} 
Instead of directly using the time-series signals,  we extract time domain and frequency domain features to better represent ECG signals. The time-domain features include: maximum, minimum, range, mean, median, mode, standard deviation, root mean square, mean square, k-order moment and skewness, kurtosis, kurtosis factor, waveform factor, pulse factor, and margin factor. The frequency-domain features include: FFT mean,  FFT variance,  FFT entropy,  FFT energy,  FFT skew,  FFT kurt,  FFT shape mean,  FFT shape std,  FFT shape skew, FFT kurt, where the function of each component is shown in Table~\ref{fre_table}.

\begin{table}[htp]\small
    \centering
    \renewcommand\arraystretch{1.5}
	\caption{ECG statistical features in frequency domain.}
		\begin{tabular}{c|c|c|c}  
			\hline
			Feature Symbol &Formula   & Feature Symbol &Formula \\  \hline
			$Z_1$  & $\frac{1}{N} \sum_{k=1}^{N} F(k)$ & $Z_6$  & $\frac{1}{N} \sum_{k=1}^{N}\left(\frac{F(k)-Z_{1}}{\sqrt{Z_{2}}}\right)^{4}$ \\
			$Z_2$  & $\frac{1}{N-1} \sum_{k=1}^{N}\left(F(k)-Z_{1}\right)^{2}$  & $Z_7$  & $\frac{\sum_{k=1}^{N}(f(k)-F(k))}{\sum_{k=1}^{N} F(k)}$ \\ 
			$Z_3$  & $-1 \times \sum_{k=1}^{N}\left(\frac{F(k)}{Z_{1} N} \log _{2} \frac{F(k)}{Z_{1} N}\right)$  & $Z_8$  & $\sqrt{\frac{\sum_{k=1}^{N}\left[\left(f(k)-Z_{6}\right)^{2} F(k)\right]}{\sum_{k=1}^{N} F(k)}}$\\  
			$Z_4$  & $\frac{1}{N} \sum_{k=1}^{N}(F(k))^{2}$ & $Z_9$ & $\frac{\sum_{k=1}^{N}\left[(f(k)-F(k))^{3} F(k)\right]}{\sum_{k=1}^{N} F(k)}$ \\  
			$Z_5$  & $\frac{1}{N} \sum_{k=1}^{N}\left(\frac{F(k)-Z_{1}}{\sqrt{Z_{2}}}\right)^{3}$  & $Z_{10}$ & $\frac{\sum_{k=1}^{N}\left[(f(k)-F(k))^{4} F(k)\right]}{\sum_{k=1}^{N} F(k)}$ \\  \hline
	\end{tabular}
	\label{fre_table}
\end{table}

After processing the ECG signals, we analyzed the statistics of the processed ECG data, and the result is shown in Table~\ref{imbalance_table}, where there are five categories in total, including NORM, MI, STTC, CD, and HYP.

\begin{table}[htp]\small
    \centering
	\caption{Statistics of the processed ECG data.}
		\begin{tabular}{c|c|c|c|c}
			\hline
			Category & Patients & Percentage & ECG beats &Percentage \\   \hline  
			NORM  &9528  &34.2\%  & 28419 &36.6\% \\ 
			MI  &5486  &19.7\%  &10959 &14.1\% \\ 
			STTC  &5250  &18.9\%   & 8906 &11.5\%   \\ 
			CD  &4907  &17.6\%  & 20955  &27.0\%  \\ 
			HYP  &2655  &9.5\%  & 8342  &10.8\%  \\ \hline
	\end{tabular}
	\label{imbalance_table}
\end{table}

\section{Experiments}
\subsection{Experimental Setup}
We use the MF-Transformer model as our classifier, where the input contains three parts: the ECG signals, the time domain features, and the frequency domain feature. To reduce the dimension of ECG signals for the convenience of computation, we downsample the processed ECG signals to 50Hz. We computed the ECG features in Section~\ref{sec:features} for each ECG beat for all 12 leads, and concatenated them with the downsampled and de-noised ECG signals. The dimension of the final features vector of each ECG beat is 864, where the dimensions for the ECG signals, time-domain features, and frequency domain features are 600, 156, and 108, respectively. Our experiments are carried out on two NVIDIA RTX A6000 GPUs.

\begin{figure}[htp]
	\centering
	\includegraphics[width=0.48\textwidth]{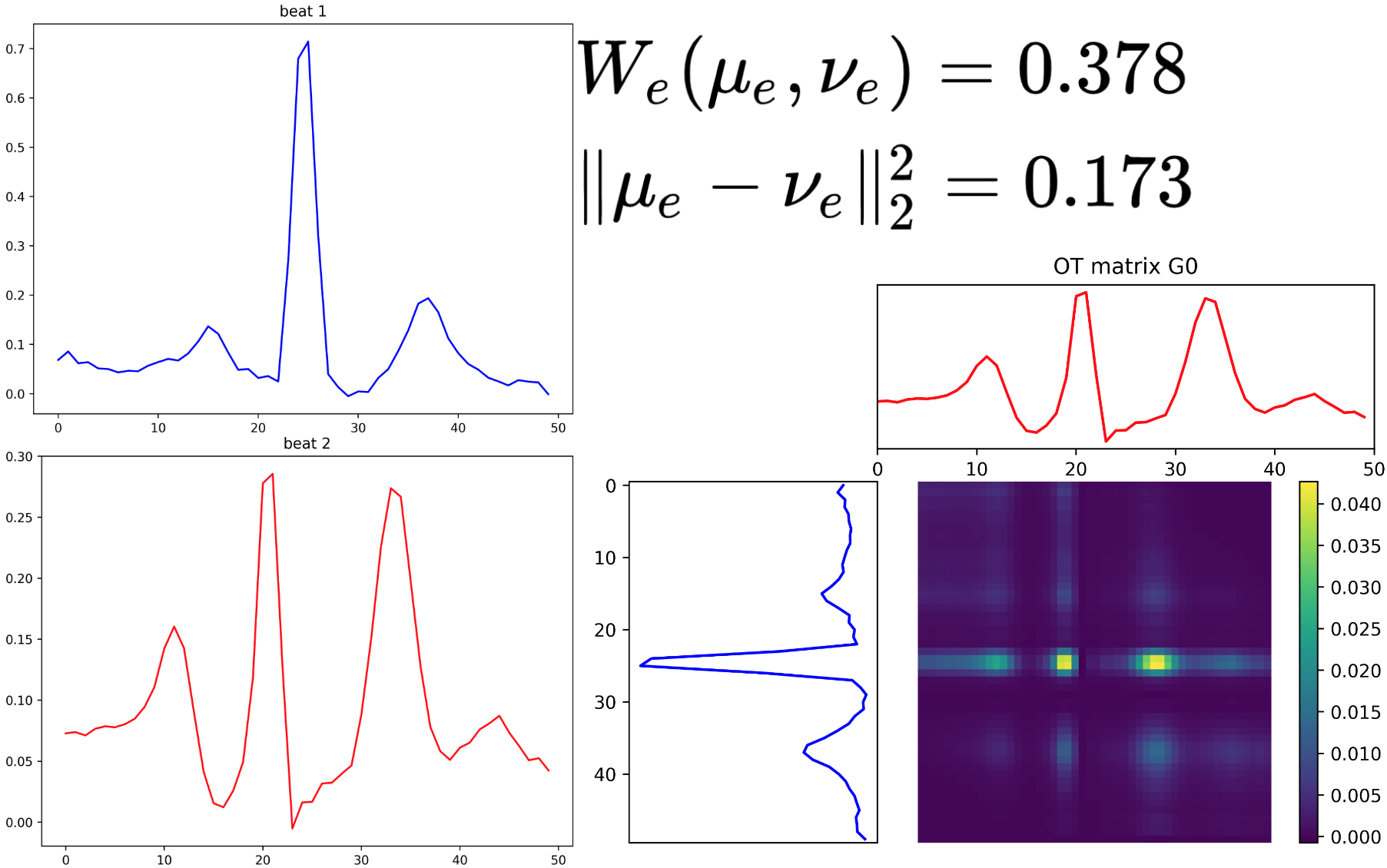}
	\includegraphics[width=0.48\textwidth]{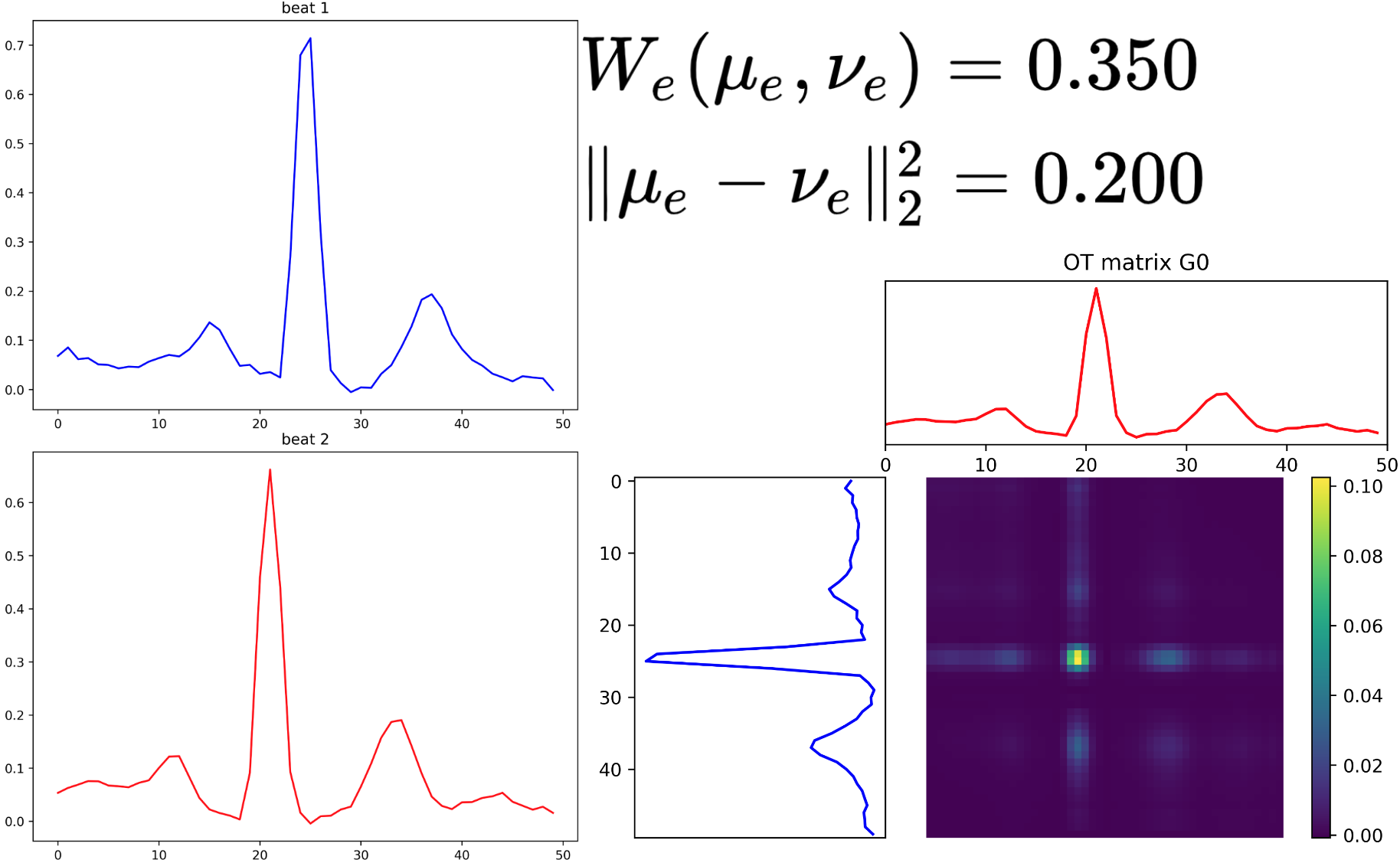}
	\caption{Difference between W-distance and l2 distance. We overlaid two ECG signals from different conditions on the left side, while on the right side, we put two ECG signals from the same condition. The Wasserstein distance correctly describes the similarity, but the $l_2$ norm indicates a large distance between two signals comes from the same condition but has a slight time shift. }
	\label{fig:geo_dist_comparision}
\end{figure}


\subsection{Data Augmentation by Wasserstein Geodesic Perturbation}
Our data augmentation strategy through Wasserstein Geodesic Perturbation aims at improving the robustness of heart disease diagnosis.
In specific, 
(1) We use NORM individual beats as the source and transport the samples from the NORM into each other minor categories. 
(2) In the augmentation procedure, we randomly sample a batch of ECG signals from both the source and target categories and then use formulation in Section~\ref{sec:DA} to get the barycentric mapping samples. The label of augmented samples is set to be the target categories.
(3) We mix the original data and augmented data together as input data for the MF-Transformer. 

\begin{figure}[htp]
	\centering
	\includegraphics[width=0.99\textwidth]{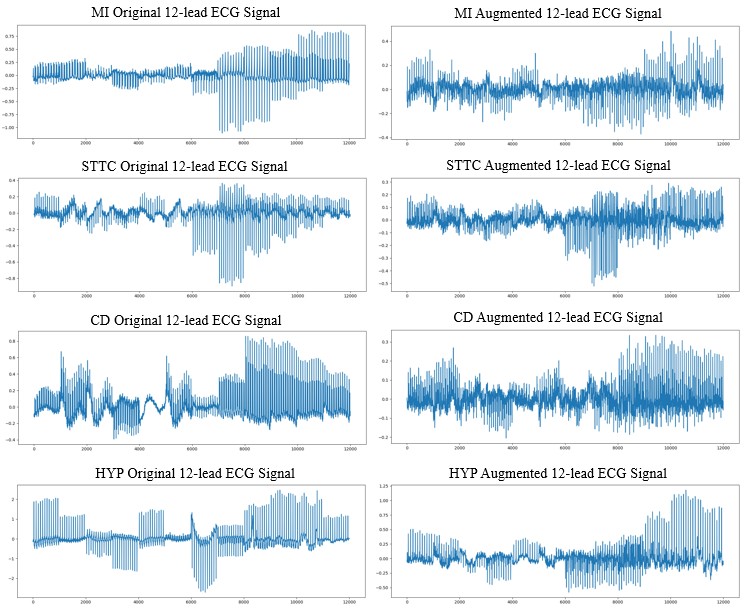}
	\caption{Example comparisons of 10-second 12-lead ECG signals within different conditions. Left column: original signals; Right column: augmented signals.}
	\label{fig:aug_result}
\end{figure}

Examples of augmented data are shown in Fig.~\ref{fig:aug_result}. The quality of the augmented data is also confirmed to preserve the semi-periodic nature. The augmentation results of each lead fit well with the ECG pattern compared with the original ECG signals.

\begin{figure}[htp]
	\centering
	\includegraphics[width=0.98\textwidth]{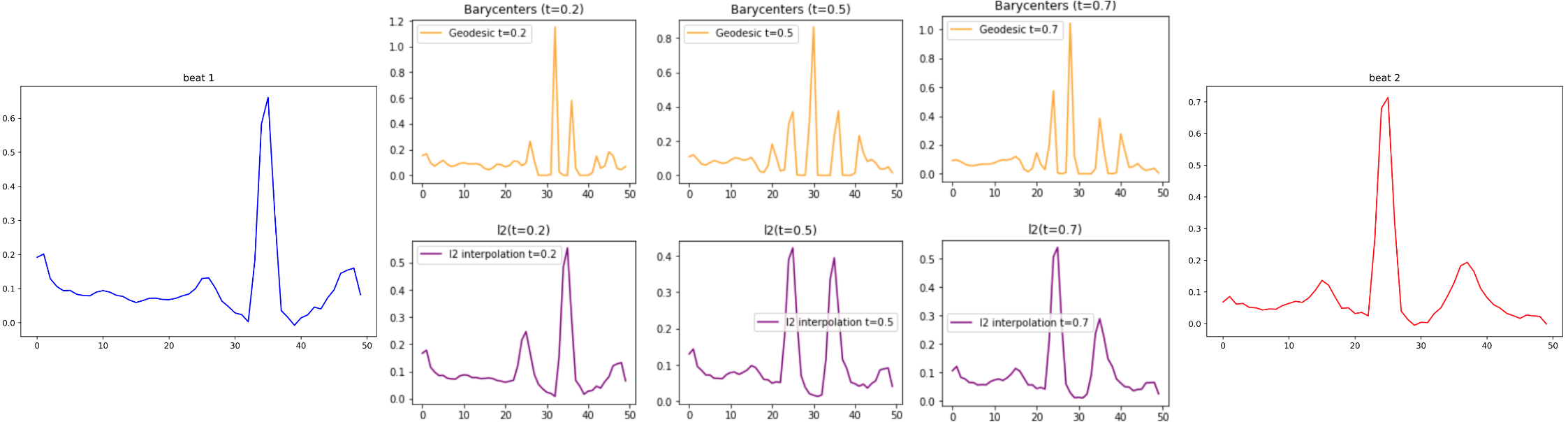}
	\caption{The difference between perturbation along the Wasserstein geodesic and on $l_2$ space. The geodesic perturbation keeps the structure of ECG beats while $l_2$ interpolation leads to situations that violates common senses (two QRS waves in one beat). }
	\label{fig:geo_inter_vs_l2}
\end{figure}

\subsection{Evaluation of Heart Disease Detection }
We used the MF-Transformer model as the classifier to evaluate our methods to detect the heart conditions based on the ECG data. First, we trained the MF-Transformer model with the original PTB-XL data to obtain the baseline performance for different categories. Second, we used different data augmentation strategies to augment the ECG signals for the minority categories, then trained the MF-Transformer model from scratch to obtain the performance by different data augmentation methods. Third, we augmented the ECG data with our data augmentation method and trained the MF-Transformer model from scratch again to evaluate the performance of our method. The augmented data is only used for training, the testing set remains the same as for all the experiments, which only contain the real-world ECG signals to have a fair evaluation of the proposed method. The training and testing splitting strategy is the same as in \citep{Wagner2020PTBXLAL,Strodthoff2021DeepLF}.  

\begin{table}[htp]\small
    \centering
	\caption{Comparison results of heart disease diagnosis (HYP, CD, STTC, MI) by different data augmentation methods, where the evaluation metrics are AUROC and F1-score.}
		\begin{tabular}{l|c|c}  
			\hline
			Methods & AUROC & F1-score \\ \hline  
			No augmentation                    &0.843  &0.575  \\  
			Random Oversampling             &0.820  &0.536   \\  
		    SMOTE \citep{Chawla2002SMOTESM} &0.799  & 0.534 \\  
			ADASYN  \citep{He2008ADASYNAS}  &0.820  &0.546    \\  
			TaskAug \citep{raghu22a}        &0.842  &--    \\ 
			Ours                            &\textbf{0.931}  &\textbf{0.707} \\ \hline
	\end{tabular}
	\label{results_table}
\end{table}

\begin{figure}[H]
	\centering
	\includegraphics[width=0.99\textwidth]{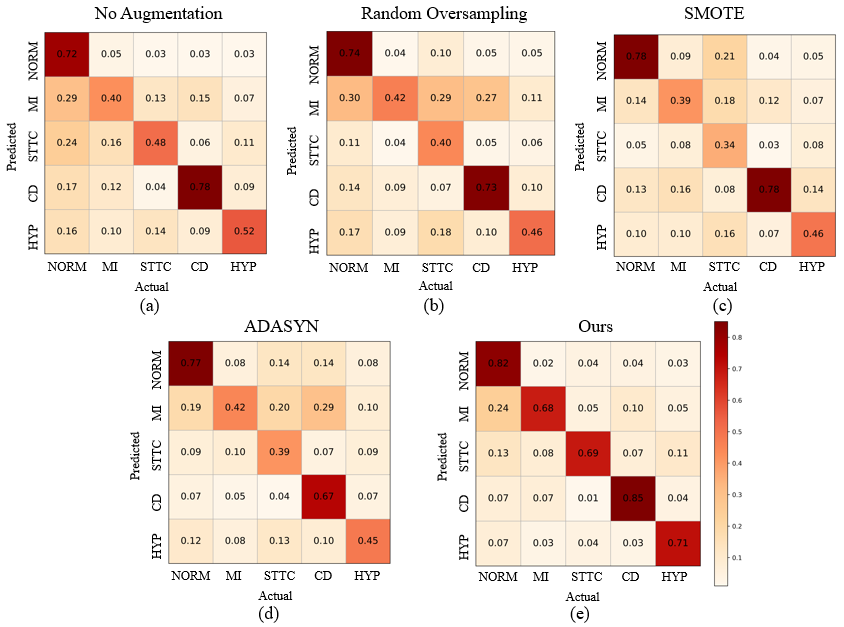}
	\caption{Confusion matrix of prediction results.}
	\label{fig:cm_all}
\end{figure}

Table~\ref{results_table} shows the results of different data augmentation approaches, where standard evaluation metrics AUROC and F1-score are used to compare the performance of different strategies. We can find that even when some data augmentation methods are applied, i.e., random oversampling, SMOTE \citep{Chawla2002SMOTESM}, or ADASYN  \citep{He2008ADASYNAS}, the classification performance is not significantly improved (or even slightly worse) than using only ECG data without any augmentation. We also compared with TaskAug \citep{raghu22a}, a new ECG data augmentation strategy which takes raw ECG signal as input.
We can find that by using no augmentation strategy, our result is already better than TaskAug, showing: (1) high-level features are useful for the diagnosis task for learning the ECG patterns; (2) some data augmentation methods' performance improvement may be due to underfitting instead of learning additional patterns. 

To have a more quantitative comparison of the classification results of each heart condition by different data augmentation methods, we compute the confusion matrix for each data augmentation method, as shown in Fig.~\ref{fig:cm_all}. Our data augmentation method not only improves the classification accuracy of each category, but also improves the average classification result. Each category's performance becomes more balanced, showing that the robustness of the diagnosis result is improved.

\paragraph{Robust prediction:} Following the pipeline of previous works~\citep{han2020deep}, we evaluate the robustness of our model as well as baseline method on the adversarial examples generated by Projected Gradient Descent (PGD)~\citep{kurakin2016adversarial_pgd}. PGD is a white-box attack methods that seeks adversarial samples with an $\epsilon$-ball based on the gradient of a trained model. In our experiment, we gradually increase the capability of the adversarial by increasing the $\epsilon$.




\begin{table}[H]\small
    \centering
	\caption{AUROC result on clean and adversarial samples, for Myocardial Infarction (MI). }
		\begin{tabular}{l|c|c|c|c|c} 
			\hline
			Myocardial Infarction (MI)  & Clean AUROC & $\epsilon=0.001$ & $\epsilon=0.002$ & $\epsilon=0.003$ & $\epsilon=0.004$  \\ \hline  
			No augmentation   &0.742   &0.563   &0.386    & 0.265 &0.185      \\ 
		    Random  Oversampling  &0.701   &0.459   &0.276    & 0.175 &0.121  \\ 
			SMOTE \citep{Chawla2002SMOTESM} & 0.706  &0.596   & 0.485   &0.403  & 0.317     \\
			ADASYN\citep{He2008ADASYNAS}& 0.726  &0.621   & 0.497   &0.405  & 0.323      \\
			Ours          &\textbf{0.910}   &\textbf{0.823}   &\textbf{0.749}    &\textbf{0.686}  &\textbf{0.615}       \\ \hline
	\end{tabular}
\end{table}
\vspace{-15pt}
\begin{table}[H]\small
    \centering
	\caption{AUROC result on clean and adversarial samples for ST/T Change (STTC). }
		\begin{tabular}{l|c|c|c|c|c} 
			\hline
			ST/T Change (STTC)  & Clean AUROC & $\epsilon=0.001$ & $\epsilon=0.002$ & $\epsilon=0.003$ & $\epsilon=0.004$  \\ \hline  
			No augmentation   & 0.847   &0.769    &0.681  &0.577  &0.481      \\ 
		    Random Oversampling  & 0.835   &0.583    &0.375  &0.247  &0.170       \\ 
			SMOTE \citep{Chawla2002SMOTESM}& 0.761   & 0.708   & 0.609 &0.524  &0.443       \\
			ADASYN \citep{He2008ADASYNAS}& 0.824   & 0.727   & 0.619 &0.525  &0.433       \\
		    Ours          &\textbf{0.935}    &\textbf{0.857}    &\textbf{0.795 } &\textbf{0.734}  &\textbf{0.680}       \\ \hline
	\end{tabular}
\end{table}
\vspace{-15pt}
\begin{table}[H]\small
    \centering
	\caption{AUROC result on clean and adversarial samples for Conduction Disturbance (CD). }
		\begin{tabular}{l|c|c|c|c|c} 
			\hline
			Conduction Disturbance (CD)  & Clean AUROC & $\epsilon=0.001$ & $\epsilon=0.002$ & $\epsilon=0.003$ & $\epsilon=0.004$  \\ \hline  
			No augmentation   & 0.883   &0.799    &0.695  & 0.603 &0.533       \\ 
		    Random Oversampling  & 0.885   &0.748    &0.598  & 0.491 &0.408      \\ 
			SMOTE \citep{Chawla2002SMOTESM}& 0.866   &0.832    & 0.786 &0.734  &0.689        \\
			ADASYN \citep{He2008ADASYNAS} & 0.869   &0.780    & 0.690 &0.615  &0.552        \\
			Ours          &\textbf{0.952}    &\textbf{0.915}    &\textbf{0.863 } &\textbf{0.811}  &\textbf{0.766 }      \\ \hline
	\end{tabular}
\end{table}
\vspace{-15pt}
\begin{table}[H]\small
    \centering
	\caption{AUROC result on clean and adversarial samples for Hypertrophy (HYP). }
		\begin{tabular}{l|c|c|c|c|c} 
			\hline
			Hypertrophy (HYP)  & Clean AUROC & $\epsilon=0.001$ & $\epsilon=0.002$ & $\epsilon=0.003$ & $\epsilon=0.004$  \\ \hline  
			No augmentation   & 0.842  &0.724   &0.599   &0.501 &0.396       \\ 
		    Random Oversampling  & 0.799  &0.588   &0.398   &0.271 &0.194       \\
			SMOTE \citep{Chawla2002SMOTESM}& 0.787  & 0.705  & 0.616  &0.532 &0.464       \\
			ADASYN \citep{He2008ADASYNAS} & 0.806  & 0.647  & 0.514  &0.419 &0.364       \\
			Ours          &\textbf{0.966}   &\textbf{0.919}   &\textbf{0.862 }  &\textbf{0.804} &\textbf{0.746}       \\ \hline
	\end{tabular}
\end{table}






\section{Conclusions, Limitation, and Future Work} 

In this paper, we propose a new method for electrocardiograms data augmentation. We perturb the dataset along the geodesic in a Wasserstein space. We show that after data augmentation, there are both accuracy and robustness improvements in the classification results over five ECG categories, which demonstrate the effectiveness of our method. Although we focus on ECG prediction in this work, our proposed data augmentation method could be applied to sequential data in other healthcare applications. 

The computational inefficiency might still be one of the significant obstacles with a large scale dataset. We would like to explore more advanced Wasserstein Barycenter algorithms that can improve efficiency. 



\acks{The work is partly supported by the Allegheny Health Network and Mario Lemieux Center for Innovation and Research in EP.}

\clearpage
\bibliography{reference}

\clearpage

\clearpage
\appendix

\section{Multi-Feature Transformer}\label{MF}
The input for the Multi-Feature Transformer is composed of three parts, including ECG raw features, time-domain features, and frequency domain features. First, we feed out the input into an embedding layer, which is a learned vector representation of each ECG feature by mapping each ECG feature to a vector with continuous values. Then we inject positional information into the embeddings by:
\begin{equation}
\begin{gathered}
P E_{(p o s, 2 i)}=\sin \left(p o s / 10000^{2 i / d_{\text {model }}}\right) \\
P E_{(p o s, 2 i+1)}=\cos \left(p o s / 10000^{2 i / d_{\text {model }}}\right)
\end{gathered}
\end{equation}
The attention model contains two sub-modules, a multi-headed attention model and a fully connected network. The multi-headed attention computes the attention weights for the input and produces an output vector with encoded information on how each feature should attend to all other features in the sequence. There are residual connections around each of the two sub-layers followed by a layer normalization, where the residual connection means adding the multi-headed attention output vector to the original positional input embedding, which helps network training by allowing gradients to flow through the networks directly. Multi-headed attention applies a self-attention mechanism, where the input goes into three distinct fully connected layers to create the query, key, and value vectors. The output of the residual connection goes through a layer normalization. 

In our model, our attention model contains $N=5$ same layers, and each layer contains two sub-layers, which are a multi-head self-attention model and a fully connected feed-forward network. Residual connection and normalization are added in each sub-layer. So the output of the sub-layer can be expressed as:
\begin{equation}
\text{Output} = \text{LayerNorm}(x+(\text{SubLayer}(x)))
\end{equation}
For the Multi-head self-attention module, the attention can be expressed as:
\begin{equation}
\text{attention} = \text{Attention}(Q,K,V)
\end{equation}
where multi-head attention uses $h$ different linear transformations to project query, key, and value, which are $Q$, $K$, and $V$, respectively, and finally concatenate different attention results: 
\begin{equation}
\text{MultiHead(Q,K,V)} = \text{Concat}(head_1, ..., head_h) W^O
\end{equation}
\begin{equation}
head_i = \text{Attention}(Q W^Q_i , K W^K_i , V W^V_i)
\end{equation}
where the projections are parameter matrices:
\begin{equation}
\begin{aligned}
&W_{i}^{Q} \in \mathbb{R}^{d_{\text {model }} d_{k}}, 
&W_{i}^{K} \in \mathbb{R}^{d_{\text {model }} d_{k}} \\
&W_{i}^{V} \in \mathbb{R}^{d_{\text {model }} d_{v}}, 
&W_{i}^{O} \in \mathbb{R}^{h d_{v} \times d_{\text {model }}}
\end{aligned}
\end{equation}
where the computation of attention adopted scaled dot-product:
\begin{equation}
\text{Attention}(Q,K,V) = \text{softmax} (\frac{Q K^T}{\sqrt{d_k}}) V
\end{equation}
For the output, we use a 1D convolutional layer and softmax layer to calculate the final output.

\section{More Related Work} 
Traditional data augmentation methods include sampling, cost-sensitive methods, kernel-based methods, active learning methods, and one-class learning or novelty detection methods \citep{He2009LearningFI}. Among them, sampling methods are mostly used, including random oversampling and undersampling,  informed undersampling, synthetic sampling with data generation, adaptive synthetic sampling, sampling with data cleaning techniques, cluster-based sampling method, and integration of sampling and boosting. But traditional methods may introduce their own set of problematic consequences that can potentially hinder learning \citep{Holte1989ConceptLA,Mease2007BoostedCT,Drummond2003C45CI}, which can cause the classifier to miss important concepts pertaining to the majority class,  or lead to overfitting \citep{Mease2007BoostedCT,He2009LearningFI}, making the classification performance on the unseen testing data generally far worse.

Optimal Transport (OT) is a field of mathematics that studies the geometry of probability spaces \citep{villani2009optimal}. The theoretical importance of OT is that it defines the Wasserstein metric between probability distributions. It reveals a canonical geometric structure with rich properties to be exploited. The earliest contribution to OT originated from Monge in the eighteenth century. Kantorovich rediscovered it under a different formalism, namely the Linear Programming formulation of OT. With the development of scalable solvers, OT is widely applied to many real-world problems \citep{Zhu2021FunctionalOT,Flamary2021POTPO}.

ECG signal can be considered as one type of sequential data, and Seq2seq models \citep{Sutskever2014SequenceTS} are widely used in time series tasks. Since the attention mechanism was proposed \citep{Bahdanau2015NeuralMT}, the Seq2seq model with attention has been improved in various tasks, which outperformed previous methods. Then Transformer model \citep{Vaswani2017AttentionIA} was proposed to solve the problem in the Seq2Seq model, replacing Long Short-Term Memory (LSTM) models with an attention structure, which achieved better results in translation tasks. The transformer model has also recently been adopted in several ECG applications, i.e., arrhythmia classification,  abnormalities detection, stress detection, etc  \citep{Yan2019FusingTM,Che2021ConstrainedTN,Natarajan2020AWA,Behinaein2021ATA,Song2021TransformerbasedSF,Weimann2021TransferLF}. But those models take only ECG temporal features as input and haven't considered the frequency domain features. To take advantage of multiple features across time and frequency domains, we proposed a Multi-Feature Transformer as our classification model to predict the heart diseases with 12-lead ECG signals.

\end{document}